\def\year{2018}\relax
\documentclass[letterpaper]{article} %DO NOT CHANGE THIS
\usepackage{aaai18}  %Required
\usepackage{times}  %Required
\usepackage{helvet}  %Required
\usepackage{courier}  %Required
\usepackage{url}  %Required
\usepackage{tabu}
\usepackage{xcolor}
\usepackage{multirow}
\usepackage{footnote}
\usepackage{graphicx}  %Required
\usepackage{subfigure}
\usepackage{amsmath,amsthm,amssymb}

\newcommand{\argmin}[1]{\underset{#1}{\operatorname{arg}\,\operatorname{min}}\;}

\begin{document}

\title{KINN: Incorporating Expert Knowledge in Neural Networks}
\author{Muhammad Ali Chattha\textsuperscript{123}, Shoaib Ahmed Siddiqui\textsuperscript{12}, Muhammad Imran Malik\textsuperscript{34},\\ {\bf \Large Ludger van Elst\textsuperscript{1}, Andreas Dengel\textsuperscript{12}, Sheraz Ahmed\textsuperscript{1}} \\
\textsuperscript{1}German Research Center for Artificial Intelligence (DFKI), Kaiserslautern, Germany.\\
\textsuperscript{2}TU Kaiserslautern, Kaiserslautern, Germany. \\
\textsuperscript{3}School of Electrical Engineering and Computer Science (SEECS),  \\National University of Sciences and Technology (NUST), Islamabad, Pakistan.\\
\textsuperscript{4}Deep Learning Laboratory, National Center of Artificial Intelligence, Islamabad, Pakistan.\\
}

\nocopyright
\maketitle

\begin{abstract}

% The ability of Artificial Neural Networks (ANNs) to learn accurate patterns from large amount of data has spurred interest of many researchers and industrialists alike. The promise of ANNs to automatically discover and extract useful features from data without dwelling on domain expertise although seems highly promising but comes at the cost of high reliance on large amount of accurately labeled data, which is often hard to acquire and formulate especially in domains like anomaly detection, natural disaster management, predictive maintenance and healthcare. We therefore propose KINN, a novel framework for incorporating expert knowledge into the network. Integrating expert knowledge into the network has two key advantages: (a) Reduction in the amount of data needed to train the model, and (b) improved convergence of model parameters. However, expert knowledge may not be comprehensive. Therefore, KINN employs a novel residual knowledge incorporation scheme which can automatically determine quality of expert predictions and adjust accordingly. Specifically, the method tries to use information contained in one domain to complement information missed by the other. We evaluated KINN on a real world traffic flow prediction problem. KINN significantly superseded performance of both the expert and as well as the base network (LSTM in this case) when evaluated in isolation, highlighting its superiority for the task.

The ability of Artificial Neural Networks (ANNs) to learn accurate patterns from large amount of data has spurred interest of many researchers and industrialists alike. The promise of ANNs to automatically discover and extract useful features/patterns from data without dwelling on domain expertise although seems highly promising but comes at the cost of high reliance on large amount of accurately labeled data, which is often hard to acquire and formulate especially in time-series domains like anomaly detection, natural disaster management, predictive maintenance and healthcare. As these networks completely rely on data and ignore a very important modality i.e. expert, they are unable to harvest any benefit from the expert knowledge, which in many cases is very useful. In this paper, we try to bridge the gap between these data driven and expert knowledge based systems by introducing a novel framework for incorporating expert knowledge into the network (KINN). Integrating expert knowledge into the network has three key advantages: (a) Reduction in the amount of data needed to train the model, (b) provision of a lower bound on the performance of the resulting classifier by obtaining the best of both worlds, and (c) improved convergence of model parameters (model converges in smaller number of epochs). Although experts are extremely good in solving different tasks, there are some trends and patterns, which are usually hidden only in the data. Therefore, KINN employs a novel residual knowledge incorporation scheme, which can automatically determine the quality of the predictions made by the expert and rectify it accordingly by learning the trends/patterns from data. Specifically, the method tries to use information contained in one modality to complement information missed by the other. We evaluated KINN on a real world traffic flow prediction problem. KINN significantly superseded performance of both the expert and as well as the base network (LSTM in this case) when evaluated in isolation, highlighting its superiority for the task.

\end{abstract}

\noindent Deep Neural Networks (DNNs) have revolutionized the domain of artificial intelligence by exhibiting incredible performance in applications ranging from image classification~\cite{krizhevsky2012imagenet}, playing board games~\cite{silver2016mastering}, natural language processing~\cite{conneau2017supervised} to speech recognition~\cite{hinton2012deep}. The biggest highlight of which was perhaps Google DeepMind's AlphaGo system, beating one of the world's best Go player, Lee Sedol in a 5 series match~\cite{wang2016does}. Consequently, the idea of superseding human performance has opened a new era of research and interest in artificial intelligence. However, the success of DNNs overshadows its limitations. Arguably the most severe limitation is its high reliance on large amount of accurately labeled data which in many applications is not available~\cite{sun2017revisiting}. This is specifically true in domains like anomaly detection, natural disaster management and healthcare. % SAS: Add references
%\textcolor{red}{Moreover, extreme data distributions can significantly hinder the performance of the network~\cite{szegedy2013intriguing}.} % TODO: Is the ref relevant?
 Moreover, training a network solely on the basis of data may result in poor performance on examples that are not or less often seen in the data and may also lead to counter intuitive results~\cite{szegedy2013intriguing}.

Humans tend to learn from examples specific to the problem, similar to DNNs, as well as from different sources of knowledge and experiences~\cite{lake2015human}. This makes it possible for humans to learn just from acquiring knowledge about the problem without even looking at the data pertaining to it. Domain experts are quite proficient in tasks belonging to their area of expertise due to their extensive knowledge and understanding of the problem, which they have acquired overtime through relevant education and experiences. Hence, they rely on their knowledge when dealing with problems. Due to their deep insights, expert predictions even serve as a baseline for measuring the performance of DNNs. Nonetheless, it can not be denied that apart from knowledge, the data also contains some useful information for solving problems. This is particularly cemented by astonishing results achieved by the DNNs that soley rely on data to find and utilize hidden features contained in the data itself~\cite{krizhevsky2012imagenet}. 

Therefore, a natural step forward is to combine both these separate streams of knowledge i.e. knowledge extracted from the data and the expert's knowledge.
%Even when lacking knowledge about any particular problem, the cognitive process of humans tries to make an educated guess using experiences acquired from other domains, that may not even be relevant to the problem in question. This enables human to share meaningful experiences across different domains while solving problems. 
 %DNNs still lacks the intuitive sense to share knowledge from heterogeneous sources. 
As a matter of fact, supplementing DNNs with expert knowledge and predictions in order to improve their performance has been actively researched upon. A way of sharing knowledge among classes in the data has been considered in zero-shot-learning \cite{rohrbach2011evaluating}, where semantic relatedness among classes is used to find classes related to the known ones. Although such techniques employ knowledge transfer, they are restricted solely to the data domain and the knowledge is extracted and shared from the data itself without any intervention from the expert. 
Similarly, expert knowledge and opinions are incorporated using distillation technique where expert network produces soft predictions that the DNN tries to emulate or in the form of posterior regularization over DNN predictions~\cite{hinton2015distilling}. %However, such techniques do not cater for a scenario where expert predictions themselves are incorrect.
% DNNs, although have achieved significant results in many applications, still lacks in imitating the cognitive process of the humans and since NNs were inspired by neurons in the human brain, it is only natural to take inspiration from cognitive process of humans to supplement NNs as well. 
All of these techniques try to strengthen DNN with expert knowledge. 
% \textcolor{red}{However, supporting the expert network is not considered to deal with scenarios where solutions are not explicitly mentioned in the expert model. }
However, cases where the expert model is unreliable or even random have not been considered.
% Moreover, directly trying to mimic expert network predictions has a natural disadvantage that strengths of DNNs are somewhat ignored and the network also becomes dependent upon quality of expert model along with quality of labeled data.
Moreover, directly trying to mimic expert network predictions has an implicit assumption regarding the high quality of the predictions made by the expert. 
We argue that the ideal incorporation of expert network would be the one where strengths of both networks are promoted and weaknesses are suppressed. Hence, we introduce a step in this direction by proposing a novel framework, Knowledge Integrated Neural Network (KINN), which aims to constructively integrate knowledge in a residual scheme residing in heterogeneous sources in the form of predictions. 
% KINN strengthens the performance of the network where predictions made by the expert and DNN aligns and in cases where they disjoin, KINN tries to support the weakness of one network with strengths of the other.
KINN's design allows it to be flexible. KINN can successfully integrate knowledge in cases where either the predictions of the expert and the DNN aligns, or are completely disjoint.
%which integrates knowledge from heterogeneous sources in the form of predictions in a residual scheme.
% Hence we introduce Knowledge Integrated Neural Network (KINN), that takes motivation from human cognition process and aims to constructively integrate useful information residing in different modalities. 
%Specifically, KINN aims to incorporate the knowledge gained from experts in the model, allowing them to make better predictions while also keeping their own fortes. 
Finding state-of-the-art DNN or expert model is not the aim here but rather, the aim is to devise a strategy that facilitates integration of expert knowledge with DNNs in a way that the final network achieves the best of both worlds.

The residual scheme employed in KINN to incorporate expert knowledge inside the network has three key advantages: (a) Significant reduction in the amount of data needed to train the model, since the network has to learn a residual function instead of learning the complete input to output space projection, (b) a lower bound on the performance of KINN based on the performance of the two subsequent classifiers achieving the best of both worlds, and (c) improvements in convergence of the model parameters as learning a residual mapping makes the optimization problem significantly easier to tackle. %However, as stated, complete reliance on expert predictions is also problematic, since they are based on generalizations which may not hold for every scenario.
% Moreover, by alleviating dependence of the network on large amount of training data, KINN does not become dependent on expert model but rather is robust enough to deal with situations where expert model does not explicitly state solutions.
Moreover, since the DNN itself is data driven, this makes KINN robust enough to deal with situations where the predictions made by the expert model are not reliable or even useless.
The rest of the paper is structured as follows: We first provide a brief overview of the work done in the direction of expert knowledge incorporation in the past. We then explain the proposed framework, KINN, in detail. After that, we present the evaluation results regarding the different experiments performed in order to prove the efficacy of KINN for the task of expert knowledge incorporation. Finally, we conclude the paper with the conclusion.

\section{Related Work}

Integrating domain knowledge and experts opinion into the network is an active area of research and even dates back to the early 90s. Knowledge-based Artificial Neural Networks (KBANN) was proposed by~\cite{towell1994knowledge}. KBANN uses knowledge in the form of propositional rule sets which are hierarchically structured.  
In addition to directly mapping inputs to outputs, the rules also state intermediate conclusions.
%In addition to directly mapping inputs to outputs, the defined rules also restrict the space of intermediate states.
The network is designed to have a one-to-one correspondence with the elements of the rule set, where neurons and the corresponding weights of their connections are specified by the rules. Apart from these rule based connections and neurons, additional neurons are also added to learn features not specified in the rule set. Similar approach has also been followed by~\cite{tran2018deep}. Although such approaches directly incorporates knowledge into the network, but they also limit the network architecture by forcing it to have strict correspondence with the rule base. As a result, this restricts the use of alternate architectures or employing network that does not directly follow the structure defined by the rule set.

\cite{hu2016harnessing} integrated expert knowledge using first order logic rules which is transferred to the network parameters through iterative knowledge distillation~\cite{hinton2015distilling}. The DNN tries to emulate soft predictions made by the expert network, instilling expert knowledge into the network parameters. Hence, the expert network acts as a teacher to the DNN i.e. the student network. 
The objective function is taken as a weighted average between imitating the soft predictions made by the teacher network and true hard label predictions. The teacher network is also updated at each iteration step with the goal of finding the best teacher network that fits the rule set while, at the same time, also staying close to the student network. In order to achieve this goal, KL-divergence between the probability distribution of the predictions made by the teacher network and softmax output layer of the student network is used as the objective function to be minimized. This acts as a constraint over model posterior. %The one-hot label encoding has no information regarding the distribution of labels. Therefore, in this distillation, the softmax layer of expert model is augmented with a temperature in order to ensure diffused probabilities. This ensures that the student network attempts to directly capture the distribution of the labels rather than the one-hot encoding, therefore, inheriting the complete label distribution from the teacher.
The proposed framework was evaluated for classification tasks and achieved superior results compared to other state-of-the-art models at that time. However, the framework strongly relies on the expert network for parametric optimization and does not cater for cases where expert knowledge is not comprehensive. 

Expert knowledge is incorporated for key phrase extraction by~\cite{gollapalli2017incorporating} where they defined label-distribution rules that dictates the probability of a word being a key phrase. For example, the rule enunciates that a noun that appears in the document as well as in the title is 90\% likely to be a key phrase and thus acts as posterior regularization providing weak supervision for the classification task. Similarly, KL-divergence between the distribution given by the rule set and the model estimates is used as the objective function to be used for the optimization. Again, as the model utilizes knowledge to strengthen the predictions of the network, it shifts the dependency of the network from the training data to accurate expert knowledge which might just be an educated guess in some cases. 
Similarly, ~\cite{xu2017semantic} incorporated symbolic knowledge into the network by deriving a semantic loss function that acts as a bridge between the network outputs and the logical constraints. The semantic loss function is based on constraints in the form of propositional logic and the probabilities computed by the network. During training, the semantic loss is added to the normal loss of the network and thus acts as a regularization term. This ensures that symbolic knowledge plays a part in updating the parameters of the network. 

\cite{wu2016knowledge} proposed a Knowledge Enhanced Hybrid Neural Network (KEHNN). KEHNN utilizes knowledge in conjunction with the network to cater for text matching in long texts. Here, knowledge is considered to be the global context such as topics, tags etc. obtained from other algorithms that extracts information from multiple sources and datasets. They employed the twitter LDA model~\cite{zhao2011comparing} as the prior knowledge which was considered useful in filtering out noise from long texts. A special gate known as the knowledge gate is added to the traditional bi-directional Gated Recurrent Units (GRU) in the model which controls how much information from the expert knowledge flows into the network.

\section{KINN: The Proposed Framework}
\subsection{Problem Formalization}
 
Time-series forecasting is of vital significance due to its high impact, specifically in domains like supply chain~\cite{fildes2015information}, demand prediction~\cite{pacchin2017comparison}, and fault prediction~\cite{baptista2018forecasting}.
% In a typical forecasting setting, the input sequence $x(t)$ along with its lagged versions are used to predict the value at time $t$. Hence the model is a functional mapping from past observations to the future value. The output of the model can be mathematically expressed by Equation\ref{eq1}
In a typical forecasting setting, a sequence of values $\{x_{t-1}, x_{t-2}, ..., x_{t-p}\}$ from the past are used to predict the value of the variable at time-step $t$, where $p$ is the number of past values leveraged for a particular prediction, which we refer as the window size. Hence, the model is a functional mapping from past observations to the future value. This parametric mapping can be written as: 
 
 \begin{equation*}
     \hat{x}_{t} = \phi([x_{t-1}, x_{t-2},..., x_{t-p}]; \mathcal{W})
 \end{equation*}
 
 \noindent where $\mathcal{W} = {\{W_l, b_l\}}_{l=1}^{L}$ encapsulates the parameters of the network and $\phi:\mathbb{R}^{p} \mapsto \mathbb{R}$ defines the map from the input space to the output space. 
 % \noindent where $\mathcal{W}$ encapsulates the parameters of the network ($\mathcal{W} = {\{W_l, b_l\}}_{l=1}^{L}$). 
%  To measure the performance Mean Squared Error (MSE) is used as the loss function to minimize, as shown in Eq.
The optimal parameters of the network $\mathcal{W}^{*}$ are computed based on the empirical risk computed over the training dataset. Using MSE as the loss function, the optimization problem can be stated as:

 \begin{equation}\label{unconditionedOptimProblem}
      \mathcal{W}^{*} = \argmin{\mathcal{W}} \frac{1}{|\mathcal{X}|} \sum_{\mathbf{x} \in \mathcal{X}} (x_{t} - \phi([x_{t-1},..., x_{t-p}]; \mathcal{W}))^2
 \end{equation}
 
\noindent where $\mathcal{X}$ denotes the set of training sequences and $\mathbf{x} \in \mathbb{R}^{p+1}$.
Solving this optimization problem, comprising of thousands if not millions of parameters, requires large amount of data to successfully constrain the parametric space so that a reliable solution is obtained.

Humans on the other hand, leverage their real-world knowledge along with their past-experiences in order to make predictions about the future.
%. Having access to this real-world knowledge significantly reduces the number of past-experiences required for the humans to make accurate predictions. % TODO: Add citation
The aim of KINN is to inject this real-world knowledge in the form of expert into the system. However, as mentioned, information from the expert may not be reliable, therefore, KINN proposes a novel residual learning framework for the incorporation of expert knowledge into the system. The residual framework conditions the prediction of the network on the expert's opinion. As a result, the network acts as a correcting entity for the values generated by the expert. This decouples our system from complete reliance on the expert knowledge.

% TODO: Replace 
%  Recurrent Neural Networks (RNNs) like Long Short Term Memory (LSTM), Gated Recurrent Unit (GRU) perform relatively well for forecasting problem. However, they suffer when dealing with unusual or anomalous circumstances. Human experts have a knack of not only following trends shown by the data but also extracting useful information from other domains that might effect future trends, for example political policies, weather etc. Hence they might be able to predict extra ordinary changes in trend due to their broader vision and sometimes, due to mere intuition. The proposed framework allows NN parameters to incorporate expert opinion in order to match human dexterity. 
 
\subsection{Dataset}
 
\begin{figure}[!t]
     \centering
     \includegraphics[width=\linewidth]{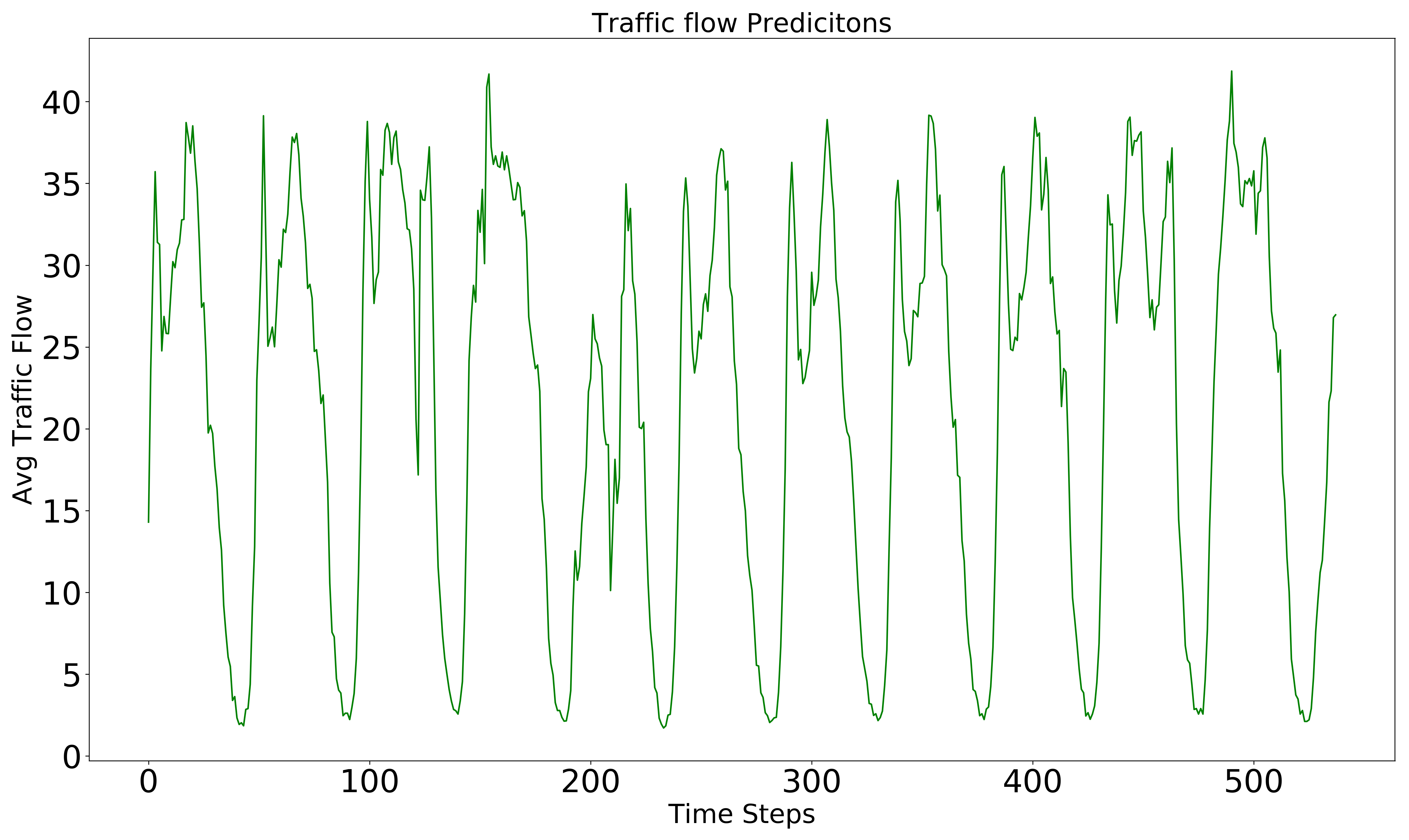}
     \caption{Traffic flow data grouped into 30 minute windows}
     \label{fig:data}
 \end{figure}

We evaluated KINN on Caltrans Performance Measurement System (PeMS) data. The data contains records of sensor readings that measure the flow of vehicular traffic on California Highways. 
%Although the complete data contained in PeMS is enormous and contains records of multiple highways, for this particular paper we only have considered the traffic flow on Richards Ave, from January 2016 till March 2016.
Since the complete PeMS dataset is enormous in terms of its size comprising of records from multiple highways, we only considered a small fraction of it for our experiments i.e. the traffic flow on Richards Ave, from January 2016 till March 2016\footnote{http://www.stat.ucdavis.edu/~clarkf/}.
The dataset contains information regarding the number of vehicles passing on the avenue every 30 seconds. PeMS also contains other details regarding the vehicles, however, we only consider the problem of average traffic flow forecasting in this paper. %prediction of average traffic flow is used as prediction problem. 
The data is grouped into 30 minute windows. The goal is to predict average number of vehicles per 30 seconds for the next 30 minutes. Fig.~\ref{fig:data} provides an overview of the grouped dataset. The data clearly exhibits a seasonal component along with high variance for the peaks. %variations in the peaks.
 
 \subsection{Baseline Expert and Deep Models}

LSTMs have achieved state-of-the-art performance in a range of different domains comprising of sequential data such as language translation~\cite{weiss2017sequence}, and handwriting and speech recognition~\cite{zhang2018drawing,chiu2018state}.
Since we are dealing with sequential data, hence, LSTM was a natural choice as our baseline neural network model. Although the aim of this work is to find a technique to fuse useful information contained in the two different modalities irrespective of their details, we nonetheless spent significant compute time to discover the optimal network hyperparameters through grid-search confined to a reasonable hyperparameter space. The hyperparameter search space included number of layers in the network, number of neurons in each layer, activation function for each layer, along with the window size $p$. 
 
\begin{figure}[!t]
    \centering
    \includegraphics[width=\linewidth]{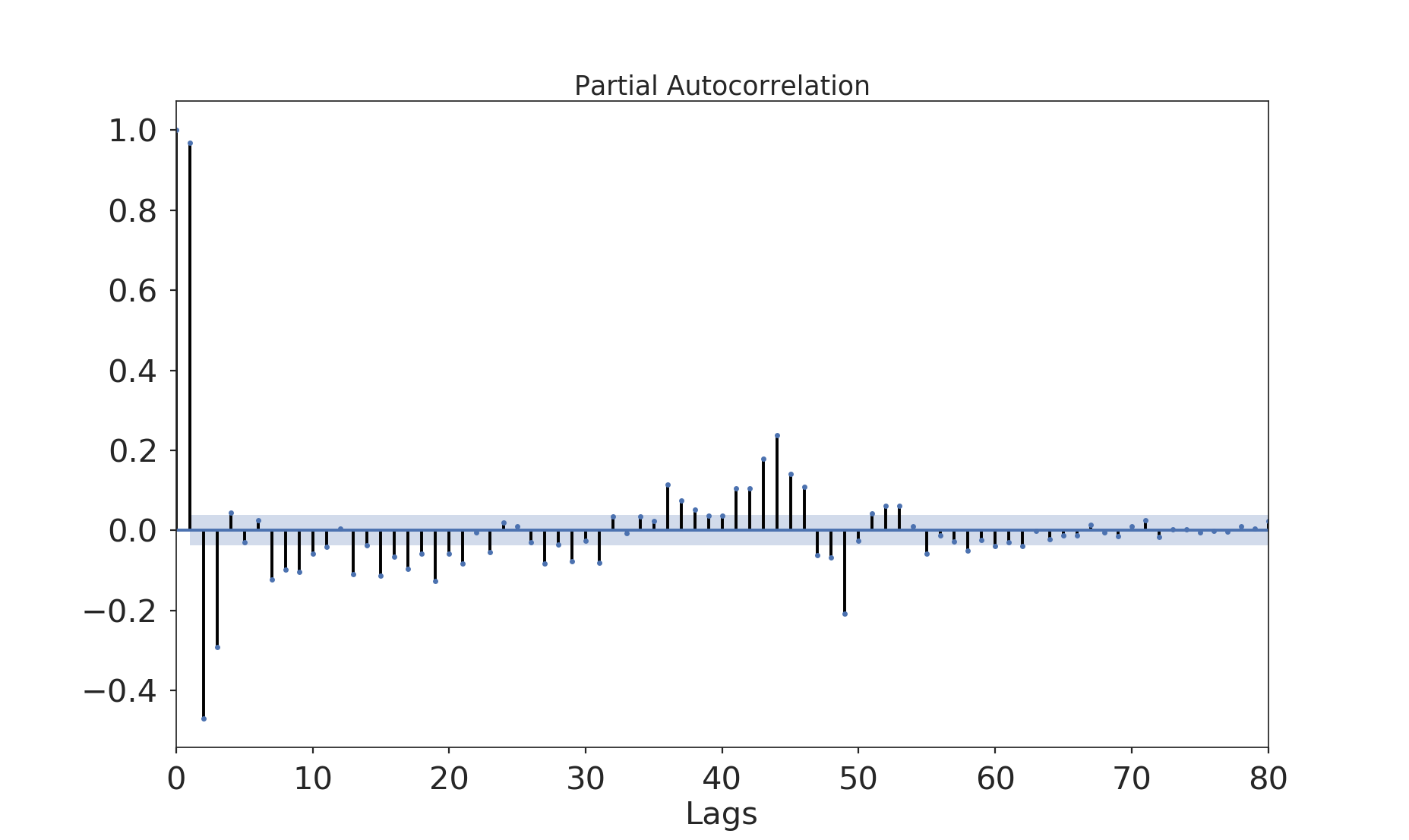}
    \caption{Partial auto-correlation of time-series}
    \label{fig:pacf}
\end{figure}

\begin{figure}[!t]
    \centering
    \includegraphics[width=1\linewidth]{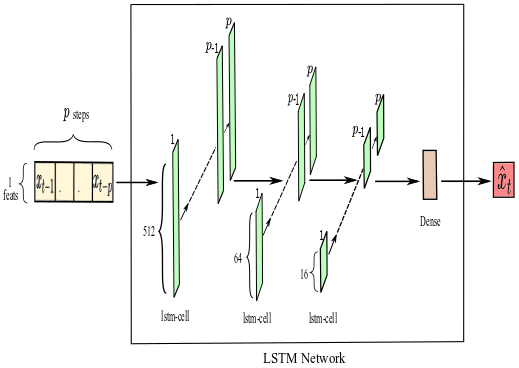}
    \caption{Neural network architecture}
    \label{fig:NN_model}
\end{figure}

% As stated, LSTMs generally perform better in time series domain due to their ability to selectively remember patterns in the series, therefore LSTM based network is used as base NN model for evaluation. 
% Although, the aim of this paper is not to find the best NN and expert model for the task but rather to integrate expert opinion in such a way that the capabilities of NN model are enhanced, still a lot of effort has been put into finding the optimal NN model. For this purpose, the base NN model is chosen after extensive grid search on model hyperparameters which included number of layers in the network, number of neurons in each layer, activation functions for each layer and the number of past values to be used for predictions i.e the window size. 
Partial auto-correlation of the series was also analyzed to identify association of the current value in the time-series with its lagged version as shown in Fig.~\ref{fig:pacf}. As evident from the figure, the series showed strong correlation with its past three values. This is also cemented by the result of the grid-search that chose the window size of three. The final network consisted of three hidden LSTM layers followed by a dense regression layer. Apart from the first layer, which used sigmoid, Rectified Linear Unit (ReLU)~\cite{glorot2011deep} was employed as the activation function. Fig.~\ref{fig:NN_model} shows the resulting network architecture. The data is segregated into train, validation and test set using 70/10/20 ratio. MSE was employed as the corresponding loss function to be optimized. The network was trained for 600 epochs and the parameters producing the best validation score were used for generating predictions on the test set.
%until the validation loss plateaus upon which the learning rate is decayed by a factor of 10 and the training is continued. The training is stopped when no further improvement in the validation loss is observed over multiple epochs.

\begin{figure*}[!t]
     \centering
     \subfigure[Predictions over the whole test set]{\includegraphics[width=0.48\textwidth]{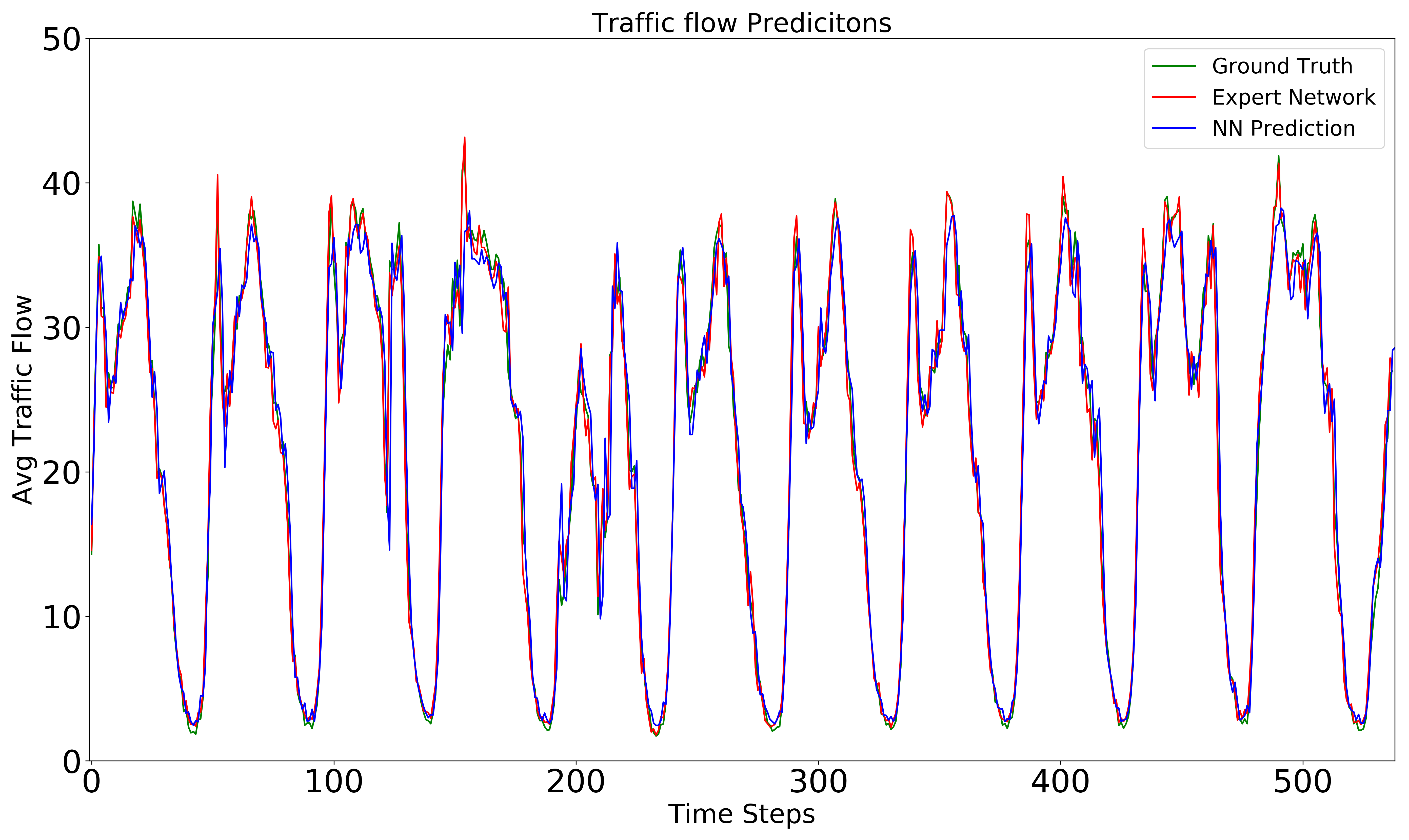}\label{fig:pred_all}}
     \subfigure[Predictions over the first 100 steps]{\includegraphics[width=0.48\textwidth]{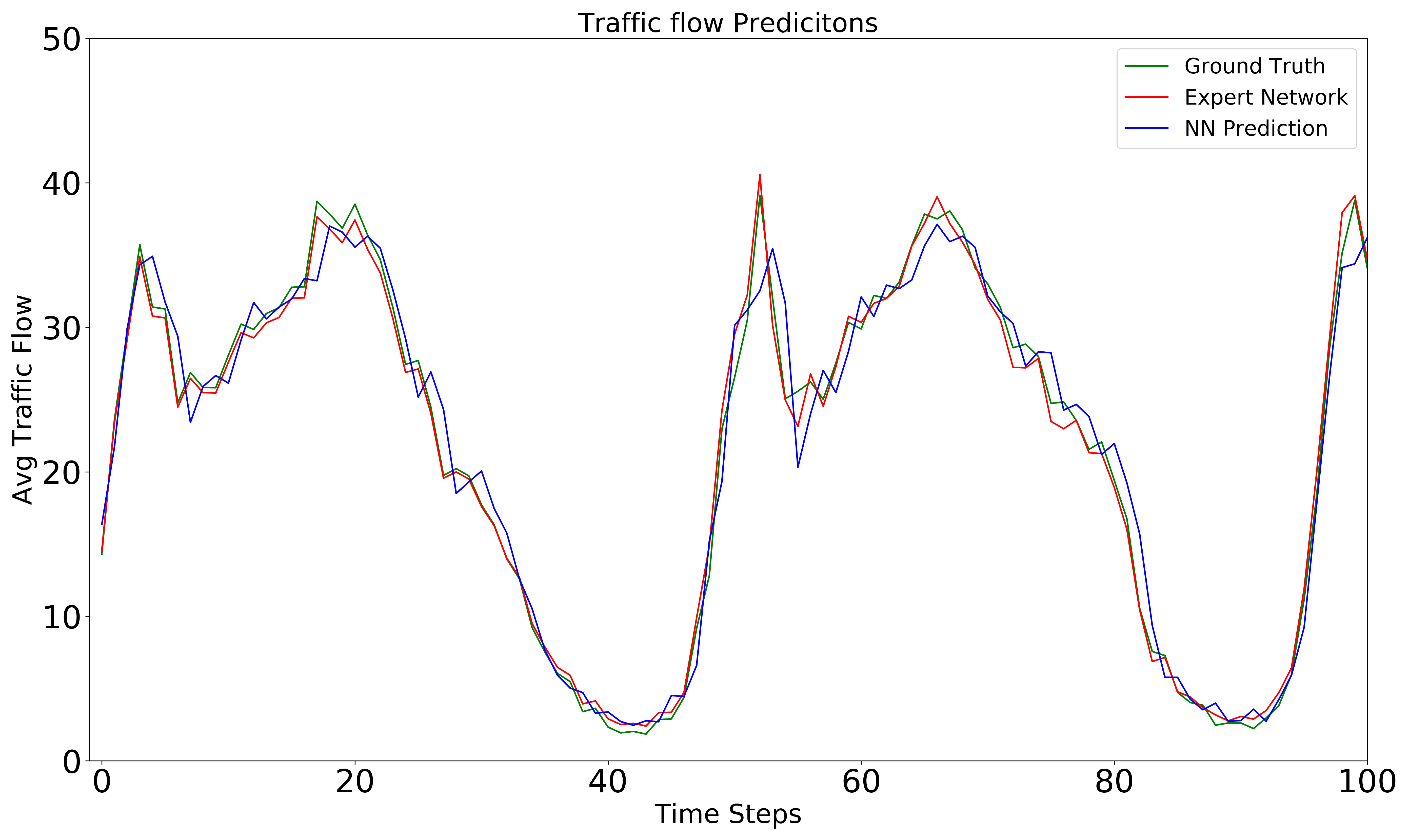}\label{fig:pred_zoom}}
     \caption{Predictions of NN and Expert Network  }
     \label{fig:pred}
 \end{figure*}

Auto-Regressive Integrated Moving Average (ARIMA) is widely used by experts in time-series modelling and analysis. Therefore, we employed ARIMA as the expert opinion in our experiments. Since the data demonstrated a significant seasonal component, the seasonal variant of ARIMA (SARIMA) was used, whose parameters were estimated using the Box-Jenkins approach~\cite{box2015time}. Fig.~\ref{fig:pred} demonstrates the predictions obtained by employing the LSTM model as well as the expert (SARIMA) model on the test set. 
 
The overall predictions made by both the LSTM as well as the expert network seems plausible as shown in Fig.~\ref{fig:pred_all}. However, it is only through thorough inspection and investigation on a narrower scale that the strengths and weaknesses of each of the networks are unveiled as shown in Fig.~\ref{fig:pred_zoom}. The LSTM tends to capture the overall trend of the data but suffered when predicting small variations in the time-series. SARIMA on the other hand was more accurate in predicting variations in the time-series. In terms of MSE, LSTM model performed considerably worse when compared to the expert model. For this dataset, the discovered LSTM model achieved a MSE of 5.90 compared to 1.24 achieved by SARIMA on the test set.

\subsection{KINN: Knowledge Integrated Neural Network}

Most of the work in the literature \cite{hu2016harnessing,gollapalli2017incorporating} on incorporating expert knowledge into the neural network focuses on training the network by forcing it to mimic the predictions made by the expert network, ergo updating weights of the network based on the expert's information. However, they do not cater for a scenario where expert network does not contain information about all possible scenarios. Moreover, these hybrid knowledge based network approaches are commonly applied to the classification scenario where output vector of the network corresponds to a probability distribution. This allows KL-divergence to be used as the objective function to be minimized in order to match predictions of the network and the expert network. In case of time-series forecasting, the output of the network is a scalar value instead of a distribution which handicaps most of the prior frameworks proposed in the literature. 
 
The KINN framework promotes both the expert model as well as the network to complement each other rather than directly mimicking the expert's output. This allows KINN to successfully tackle cases where predictions from the expert are not reliable.
Finding the best expert or neural network is not the focus here but instead, the focus is to incorporate expert prediction, may it be flawed, in such a way that the neural network maintains its strengths while incorporating strengths of the expert network.

% The challenge in hybrid, knowledge enhanced, Neural Network lies in using information contained in a one modality to offset information missing from the other. In order to achieve that, we propose modification to classic NN architectures. Firstly, instead of relying solely on the past observed value the expert prediction is also augmented in the input and the intermediary hidden layers of the NN. Secondly, expert prediction is also added to the output of final prediction layer of the NN. This forces the network to learn an underlying mapping function that is different from the one learned by traditional NN architecture, given by Eq.~\ref{eq1}. The output of the proposed framework can be expressed by Eq.~\ref{eq3}.

There are many different ways through which knowledge between an expert and the network can be integrated. Let $\hat{x}^{p}_{t} \in \mathbb{R}$ be the prediction made by the expert. We incorporate the knowledge from the expert in a residual scheme inspired by the idea of ResNet curated by~\cite{he2016deep}. Let $\phi: \mathbb{R}^{p+1} \mapsto \mathbb{R}$ define the mapping from the input space to the output space. The learning problem from Eq.~\ref{unconditionedOptimProblem} after availability of the expert information can be now be written as:

% \begin{equation}\label{eq3}
% \begin{split}
% \hat{x}_t & = \phi([x_{t-1},... ,x_{t-p}, \hat{x}^{p}_{t}]; \mathcal{W}) + \hat{x}^{p}_{t}\\
% & = \Delta + \hat{x}^{p}_{t}
% \end{split}
% \end{equation}

\begin{equation*}
     \hat{x}_{t} = \phi([x_{t-1}, x_{t-2},..., x_{t-p}, \hat{x}^{p}_{t}]; \mathcal{W}) + \hat{x}^{p}_{t}
 \end{equation*}

\begin{equation}\label{conditionedOptimProblem}
\begin{aligned}
      \mathcal{W}^{*} = \argmin{\mathcal{W}} \frac{1}{|\mathcal{X}|} \sum_{\mathbf{x} \in \mathcal{X}} (x_{t} - (\phi([x_{t-1},..., x_{t-p}, \hat{x}^{p}_{t}]; \mathcal{W}) \\+ \hat{x}^{p}_{t}))^2
\end{aligned}
\end{equation}

\begin{figure*}[!t]
    \centering
    \includegraphics[width=\linewidth]{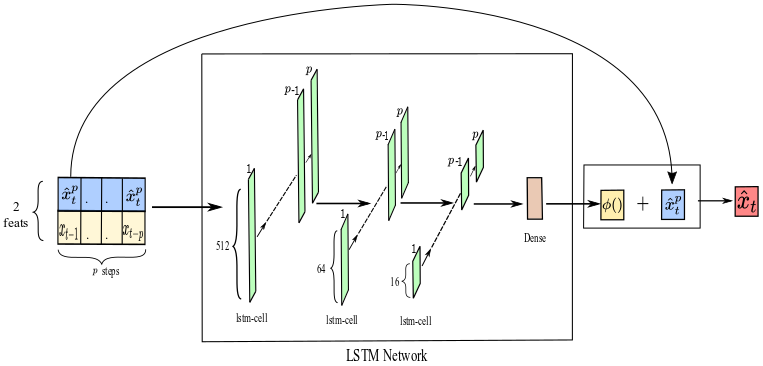}
    \caption{Proposed Architecture}
    \label{fig:my_model}
\end{figure*}

\noindent Instead of computing a full input space to output space transform as in Eq.~\ref{unconditionedOptimProblem}, the network instead learns a residual function. This residual function can be considered as a correction term to the prediction made by the expert model. Since the model is learning a correction term for the expert's prediction, it is essential for the model prediction to be conditioned on the expert's prediction as indicated in Eq.~\ref{conditionedOptimProblem}. There are two simple ways to achieve this conditioning for the LSTM network. The first one is to append the prediction at the end of the sequence as indicated in the equation. Another possibility is to stack a new channel to the input with repeated values for the expert's prediction. The second case makes the optimization problem easier as the network has direct access to the expert's prediction at every time-step. Therefore, results in minor improvements in terms of MSE. The system architecture for KINN is shown in Fig.~\ref{fig:my_model}.

Incorporating expert knowledge in this residual fashion serves a very important purpose in our case. In cases where the expert's predictions are inaccurate, the network can generate large offsets in order to compensate for the error while the network can essentially output zero in cases where the expert's predictions are extremely accurate. With this flexibility built into the system, the system can itself decide its reliance on the expert's predictions. 

% The mapping function learned by the NN is the delta function, denoted by $\Delta$ in equation \ref{eq3}, instead of the prediction function as in case of traditional NNs. We believe this is pivotal in intelligently incorporating expert knowledge. The proposed framework has the flexibility to mimic predictions made by the expert network by learning a null delta function if the data does not contain any useful information and is also versatile enough to learn offsets for expert prediction if predictions made by the expert are flawed. Alternatively, it can be said that the network tries to offset the expert prediction where it deems suitable. This allows the proposed framework to surpass performance of even the expert and hence produce outputs that are better than any of the individual network. Mathematically it looks similar to residual connections in Resnet architecture \cite{he2016deep} however there are subtle differences. For example, the skip connection comes from an entirely different domain and only includes predictions made by the expert network instead of connecting the input of the layer to its output. 

% Connections already removed
% Information contained in the expert network is also made available to the intermediate layers of the NN. The expert predictions are first projected into equivalent subspace of the hidden layer before being concatenated into the input of the hidden layer. Dense layers are used for projecting expert predictions into corresponding latent space and the process is done for every hidden layer in the NN. This gives network the freedom to use information contained in expert knowledge in any of its latent subspace. Motivation of this is drawn from Densenet\cite{huang2017densely} where feature maps from previous layers are used as input to subsequent layers however here, instead of feature maps from the previous layer, projection of expert knowledge is used. This allows knowledge incorporation into more abstract feature of NN. The proposed framework not only facilitates learning complex patterns from limited data, but also allows for faster convergence of network parameters. Fig \ref{fig:my_model} show the proposed architecture for knowledge incorporation.

\section{Evaluation}

%% SAS
% 1. Split different experiments into different sub-sections.
% 2. Add better transitions
% 3. Figure with demonstration of the conditioning (channel stacking with expert's prediction)
% 4. Correction of the results reported (removal of *)

\begin{table*}
\centering
\tabulinesep=3px
    %\begin{tabu} {|X[c]|X[c]|X[c]|X[c]|X[c]|}
    \begin{tabu} {|c|c|c|c|c|c|}
         \hline
         & & & \multicolumn{3}{c|}{MSE} \\
        \hline         
         \textbf{Experiment} & \textbf{Description} & \textbf{\% of training data used} & \textbf{DNN} & \textbf{Expert Network} & \textbf{KINN}  \\
         \hline
         \hline
         1 & Full training set and accurate expert & 100 & 5.90 & 1.24 & \textbf{0.74} \\
         \hline
         2 & Reduced training set (50\%) and accurate expert & 50 & 6.36 & 1.52 & \textbf{0.89}\\
        %  \hline
        \cline{2-6}
          & Reduced training set (10\%) and accurate expert & 10 & 6.68& 2.67 & \textbf{1.53} \\
         \hline
         3 & Full training set and noisy expert & 100 & 5.90 & 7.81 & \textbf{3.09} \\
         \hline
         4 & Reduced training set and noisy expert & 10 & 6.68 & 7.81 & \textbf{3.73} \\
         \hline
         %5 & 100 & 5.90 & 621.00 (Zero expert pred.) & \textbf{5.92} \\
         5 & Full training set and Zero expert pred. & 100 & 5.90 & 621.00 & \textbf{5.92} \\
        %  \hline
        \cline{2-6}
          & Full training set and Delayed expert pred. & 100 & 5.90 & 9.04 & \textbf{5.91} \\
         \hline
    \end{tabu}
\caption{ MSE on the test set for the experiments performed }
\label{table:1}
\end{table*}

We curated a range of different experiments each employing KINN in a unique scenario in order to evaluate its performance under varied conditions. We compare KINN results with the expert as well as the DNN in terms of performance to highlight the gains achieved by employing the residual learning scheme. To ensure a fair comparison, all of the preprocessing and LSTM hyperparameters were kept the same when the model was tested in isolation and when integrated as the residual function in KINN. 

In the first setting, we tested and compared KINN's performance in the normal case where the expert predictions are accurate and the LSTM is trained on the complete training set available. We present the results from this normal case in experiment \# 01. 
% In order to evaluate KINN's performance even in cases where the amount of training data available is small or the expert is inaccurate. In order to evaluate those claims, we spawned two new experiments from the first one. In the first case, we reduced the amount of training data provided to the models for training. We present the findings from this experiment in section~\ref{sec:accExpReducedData}.
In order to evaluate KINN's performance in cases where the amount of training data available is small or the expert is inaccurate, we established two different sets of experiments starting from the configuration employed in the first experiment. In the first case, we reduced the amount of training data provided to the models for training. We present the findings from this experiment in experiment \# 02. 
In the second case, we reduced the reliability of the expert predictions by injecting random noise. The results from this experiment are summarized in experiment \# 03. 
A direct extension of the last two experiments is to evaluate KINN's performance in cases where both of these conditions hold i.e. the amount of training data is reduced as well as the expert is noisy. We summarize the results for this experiment in experiment \# 04.
Finally, we evaluated KINN's performance in cases where the expert contained no information. We achieved this using two different ways. We first evaluated the case where the expert always predicted the value of zero. In this case, the target was to evaluate the impact (if any) of introducing the residual learning scheme since the amount of information presented to the LSTM network was exactly the same as the isolated LSTM model in the first experiment. We then tested a more realistic scenario, where the expert model replicated the values from the last time-step of the series. We elaborate the findings from this experiment (for both settings) in experiment \# 05.

\subsection{Experiment \# 01: Full training set and accurate expert} \label{sec:accExpFullData}

\begin{figure*}[!t]
    \centering
    \subfigure[Predictions of all models]{\includegraphics[width=0.48\linewidth]{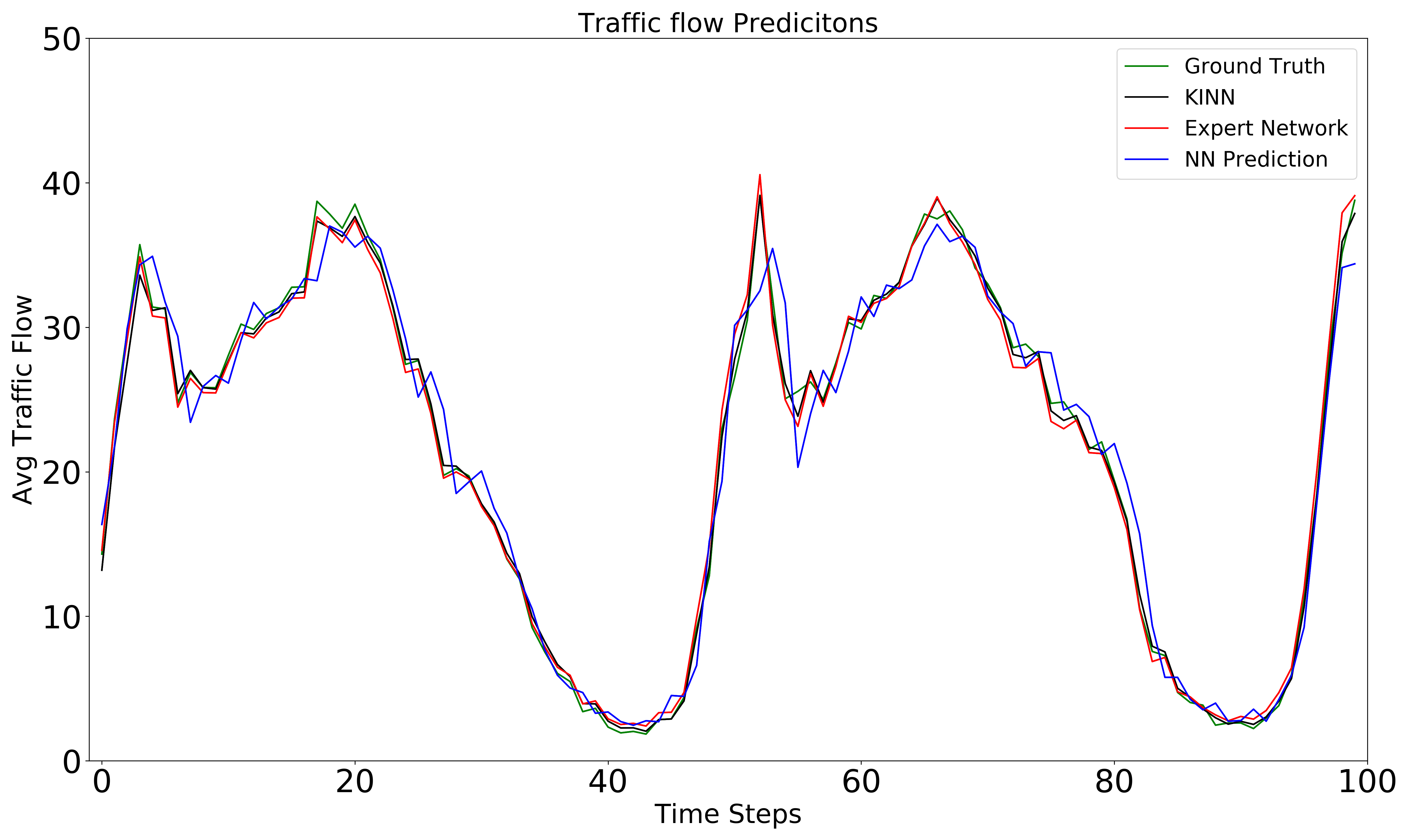}}
    \subfigure[Step-wise error plot]{\includegraphics[width=0.48\linewidth]{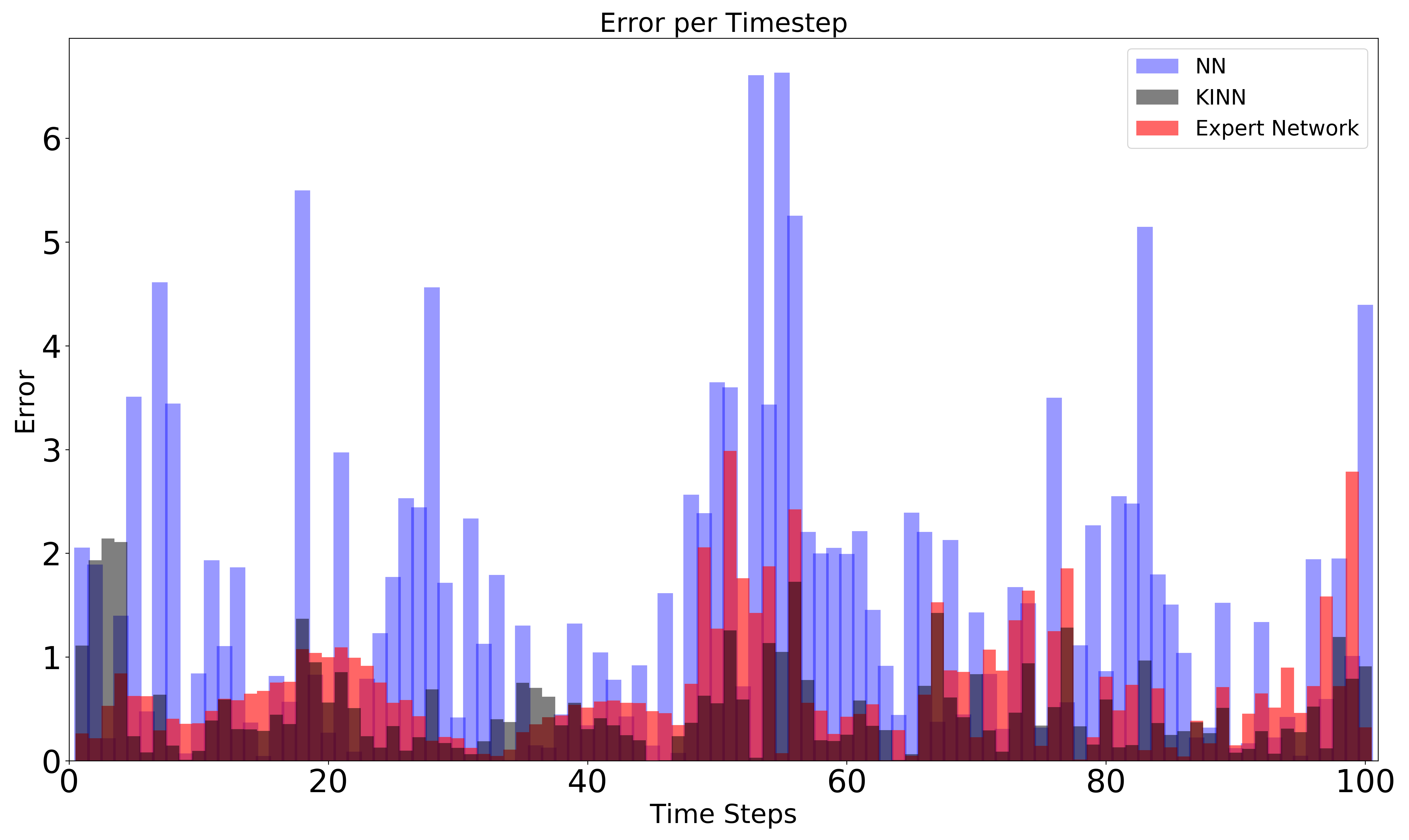}}
    \caption{Predictions and the corresponding error plot for the normal case (experiment \# 01)}
    \label{fig:normal_plot}
\end{figure*}

We first tested both the LSTM as well as the expert model in isolation in order to precisely capture the impact of introducing the residual learning scheme.
% From the start of training there is clear improvement in training of the network. 
KINN demonstrated significant improvements in training dynamics directly from the start.
KINN converged faster as compared to the isolated LSTM. As opposed to the isolated LSTM which required more training time (epochs) to converge, KINN normally converged in only one fouth of the epochs taken by the isolated LSTM, which is a significant improvement in terms of the compute time. Apart from the compute time, KINN achieved a MSE of 0.74 on the test set. This is a very significant improvement in comparison to the isolated LSTM model that had a MSE of 5.90. Even compared to the expert model, KINN demonstrated a relative improvement of 40\% in terms of MSE. Fig.~\ref{fig:normal_plot} showcases the predictions made by KINN along with the isolated LSTM and the expert network on the test set. It is evident from the figure that KINN caters for the weaknesses of each of the two models involved using the information contained in the other. The resulting predictions are more accurate than the expert network on minimas and also captures the small variations in the series which were missed by the LSTM network. 
% The results verify that the modifications taken into account during design process of the network are successful in complementing weakness of both domains.

% \begin{figure}[!t]
%     \centering
%     \includegraphics[width=\linewidth]{proposed_pred.png}
%     \caption{Predictions of KINN along with the isolated model predictions}
%     \label{fig:prop_pred}
% \end{figure}

% \begin{figure}[!t]
%     \centering
%     \includegraphics[width=\linewidth]{error_plot_100.png}
%     \caption{Step-wise error plot}
%     \label{fig:error_plot}
% \end{figure}

In order to further evaluate the results, error at each time-step is compared for the isolated models along with KINN. To aid the visualization, step-wise error for first 100 time-steps of the test set is shown in Fig.~\ref{fig:normal_plot}. The plot shows that the step-wise prediction error of KINN is less than both the expert model as well as the LSTM for major portion of the time. 

However, there are instances where predictions made by KINN are slightly worse than those of the baseline models. In particular, the prediction error of KINN exceeded the error of the expert network for only 30\% of the time-steps and only 22\% of the time-steps in case of the LSTM network. Nevertheless, even in those instances, the performance of KINN was still on par with the other models since on 99\% of the time-steps, the difference in error is less than 1.5.    

\subsection{Experiment \# 02: Reduced training set and accurate expert}\label{sec:accExpReducedData}

\begin{figure*}[!t]
    \centering
    \subfigure[Predictions of all models]{\includegraphics[width=0.48\linewidth]{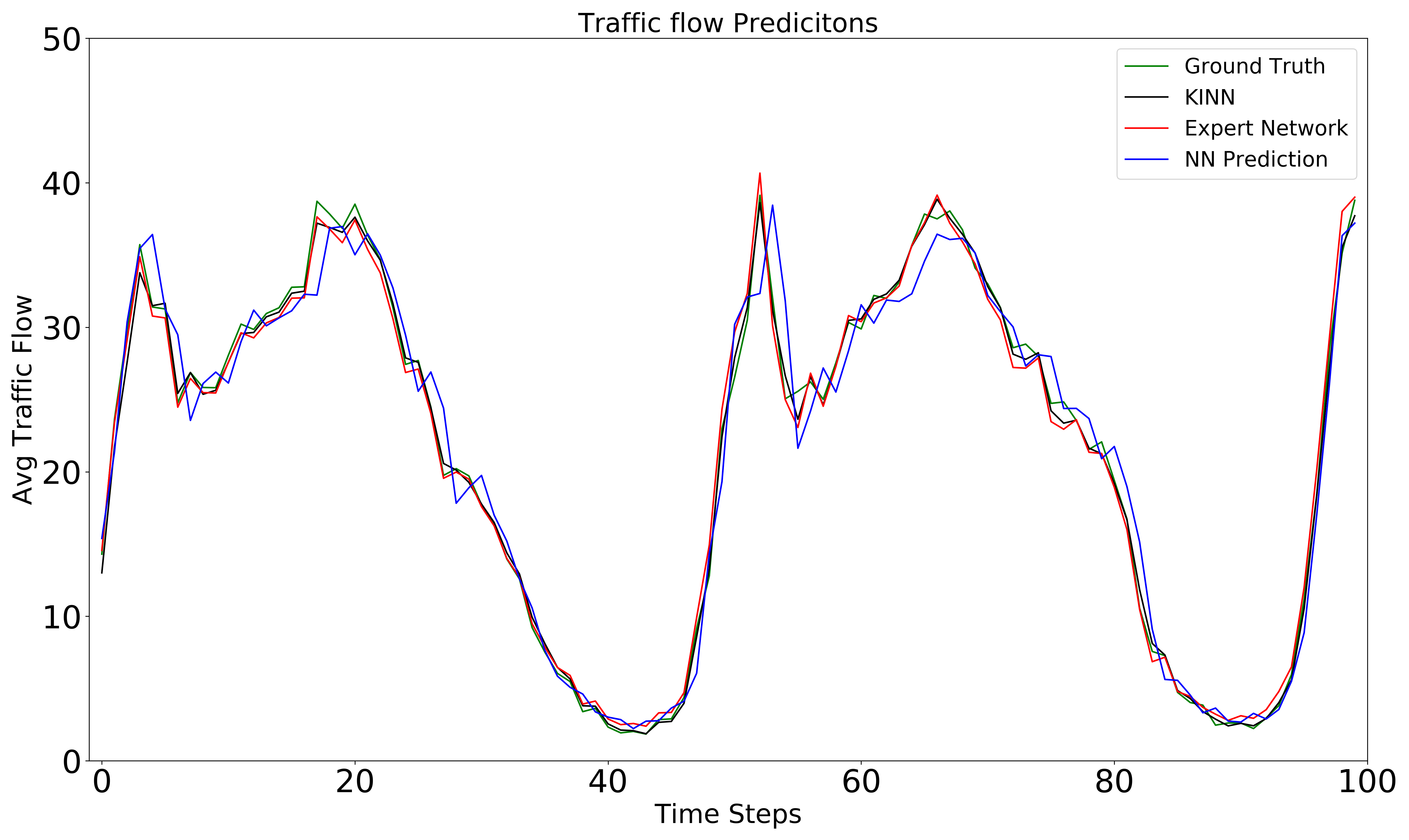}}
    \subfigure[Step wise error plot]{\includegraphics[width=0.48\linewidth]{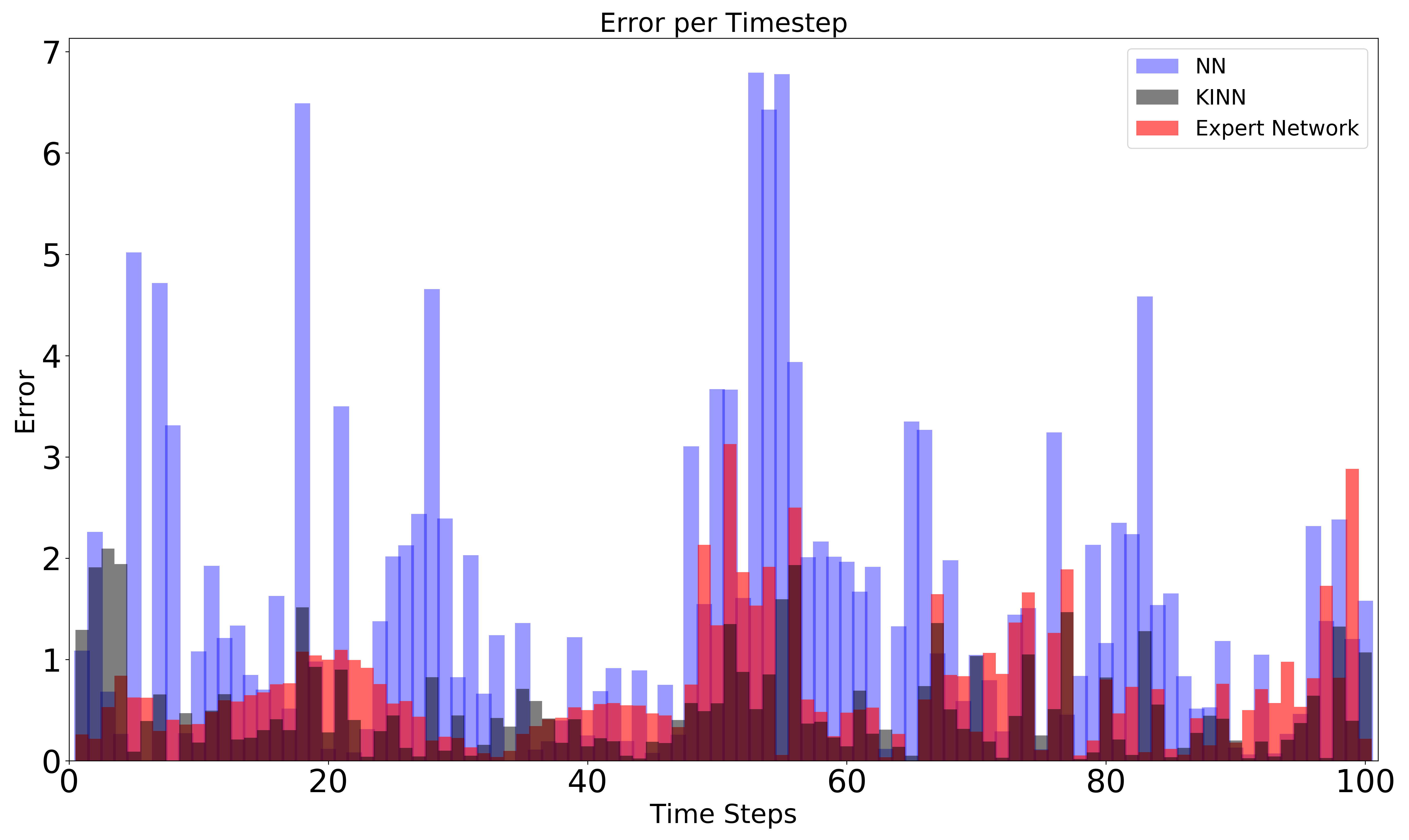}}
    \caption{Prediction and error plot with only 50\% of the training data being utilized}
    \label{fig:50_plot}
\end{figure*}

\begin{figure*}[!t]
    \centering
    \subfigure[Predictions of all models]{\includegraphics[width=0.48\textwidth]{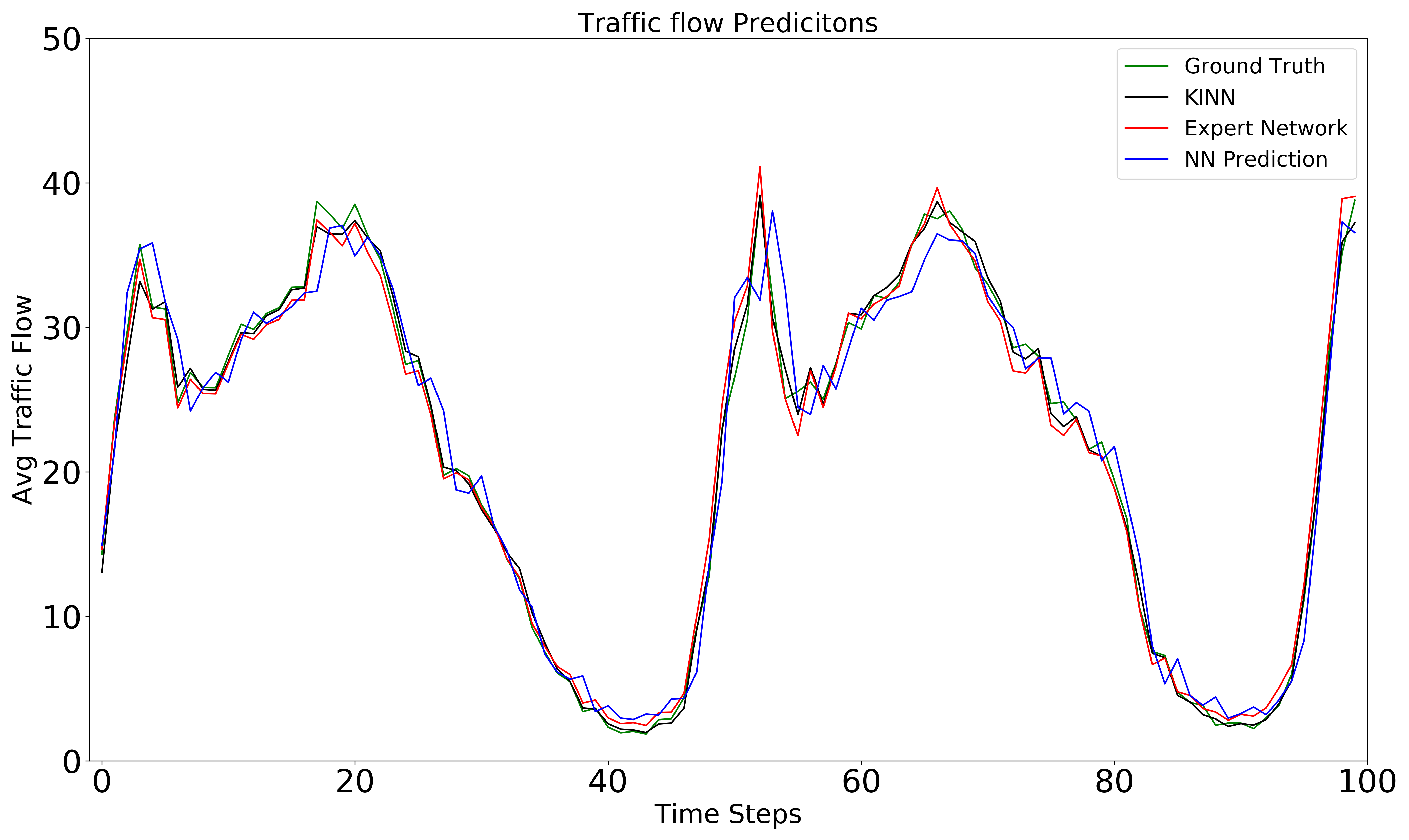}\label{subfig:10_pred}}
    \subfigure[Step wise error plot]{\includegraphics[width=0.48\textwidth]{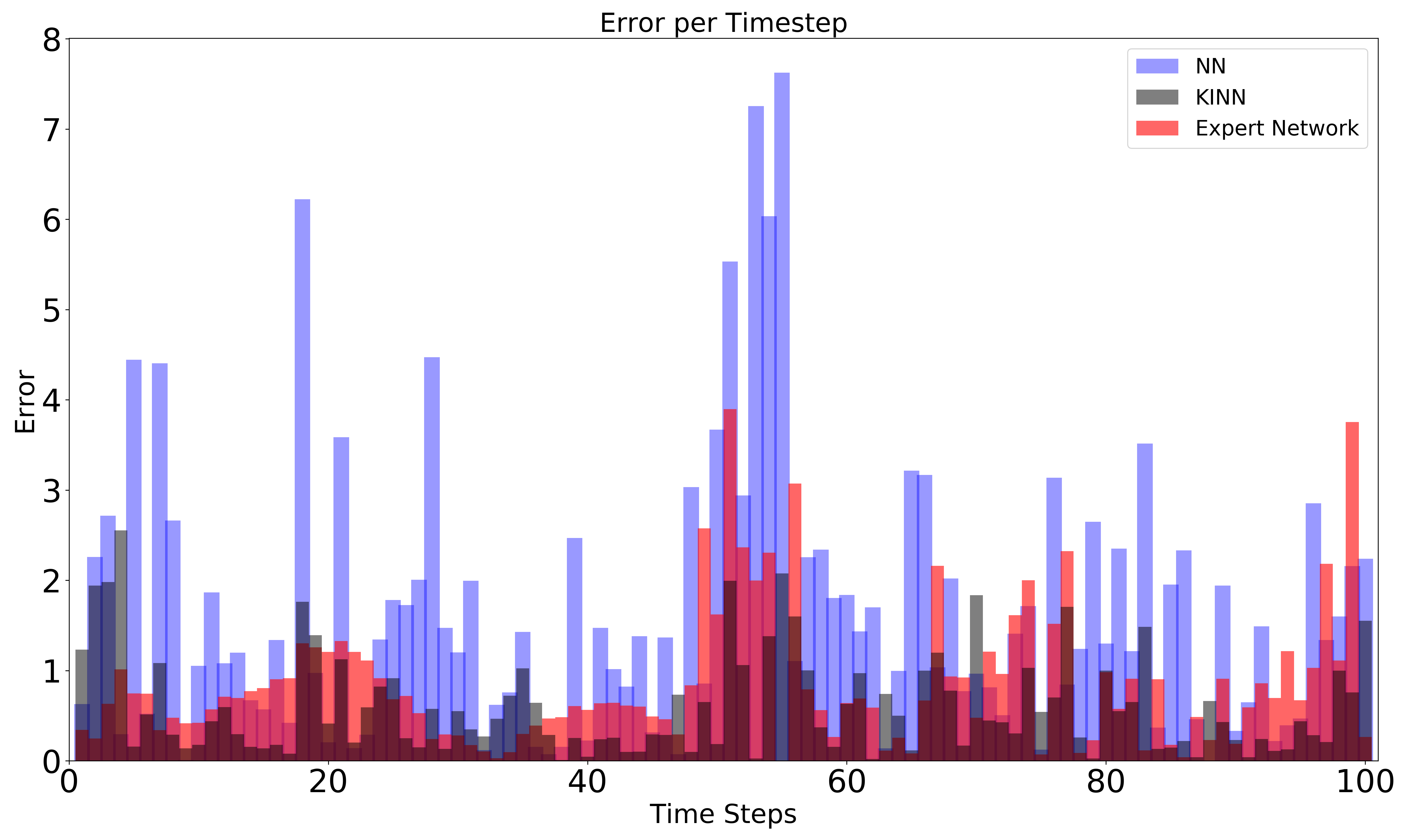}}
    \caption{Prediction and error plot with only 10\% of the training data being utilized}
    \label{fig:10_plot}
\end{figure*}

One of the objectives of KINN was to reduce dependency of the network on large amount of labelled data. 
We argue that the proposed model not only utilizes expert knowledge to cater for shortcomings of the network, but also helps in significantly reducing its dependency on the data. To further evaluate this claim, a series of experiments were performed. KINN was trained again from scratch using only 50\% of the data in the training set. The test set remained unchanged. Similarly, the LSTM network was also trained with the same 50\% subset of the training set.

The LSTM network trained on the 50\% subset of the training data attained a MSE of 6.36 which is slightly worse than the MSE of network trained on the whole training set. 
% Performance of the expert network also degraded slightly and it achieves MSE of 2.16.
Minor degradation was also observed in the performance of the expert network which achieved a MSE of 1.52.
Despite of this reduction in the dataset size, KINN achieved significantly better results compared to both the LSTM as well as the expert model achieving a MSE of 0.89. Fig~\ref{fig:50_plot} visualizes the corresponding prediction and error plots of the models trained on 50\% subset of the training data. 

We performed the same experiment again with a very drastic reduction in the training dataset size by using only 10\% subset of the training data. Fig.~\ref{fig:10_plot} visualizes the results from this experiment in the same way, by first plotting the predictions from the models along with the error plot. 
It is interesting to note that since the LSTM performed considerably poor due to extremely small training set size, the network shifted its focus to the predictions of the expert network and made only minor corrections to it as evident from Fig.~\ref{subfig:10_pred}. This highlights KINN's ability to decide its reliance on the expert predictions based on the quality of the information. In terms of the MSE, LSTM model performed the worst. When trained on only the 10\% subset of the training set, the LSTM model attained a MSE of 6.68, whereas the expert model achieved MSE of 2.67. KINN on the other hand, still outperformed both of these models and achieved a MSE of 1.53. 

\subsection{Experiment \# 03: Full training set and noisy expert} \label{sec:noisyExpFullData}

\begin{figure*}[!t]
    \centering
    \subfigure[Predictions of all models]{\includegraphics[width=0.48\linewidth]{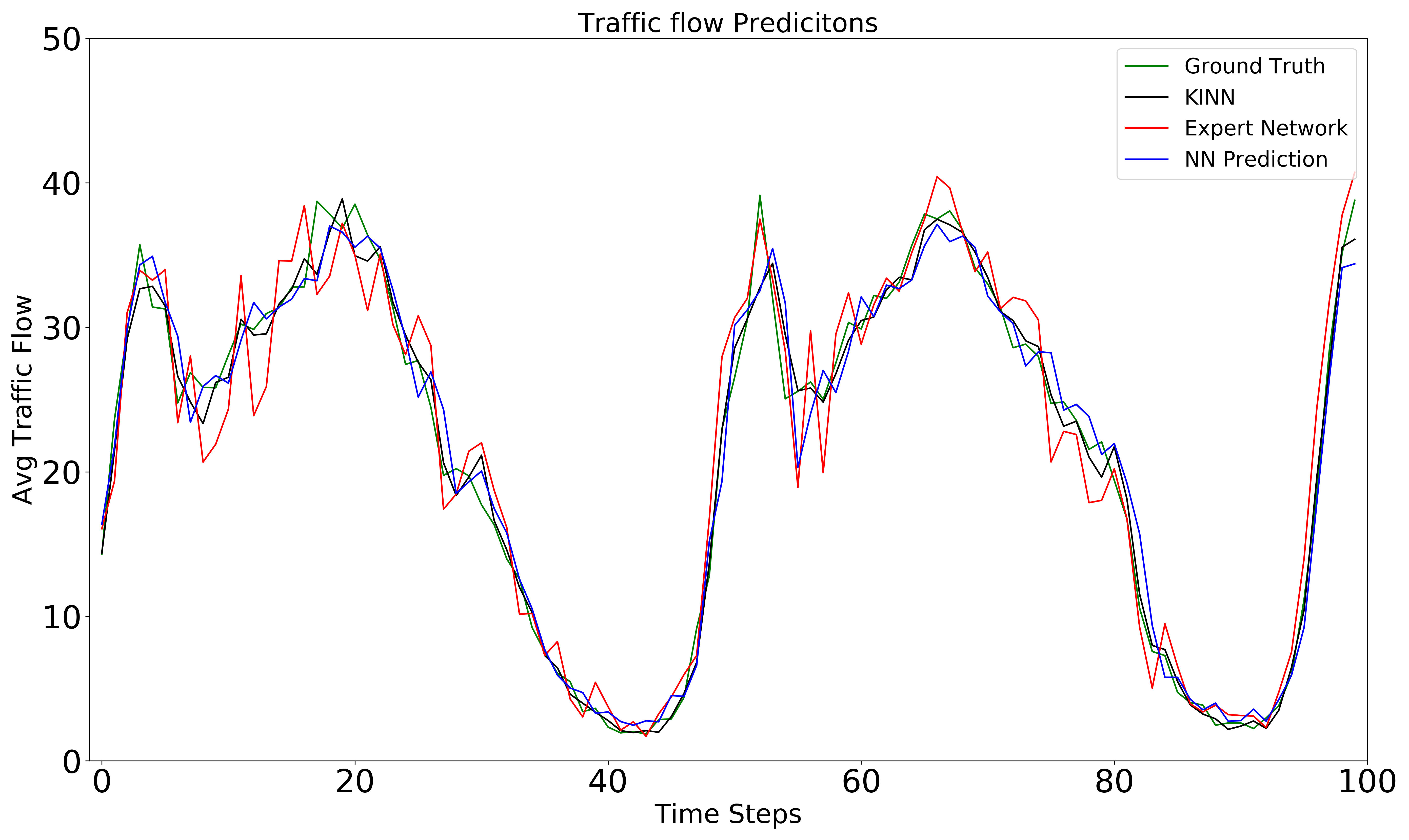}\label{fig:noisy_pred}}
    \subfigure[Step wise error plot]{\includegraphics[width=0.48\linewidth]{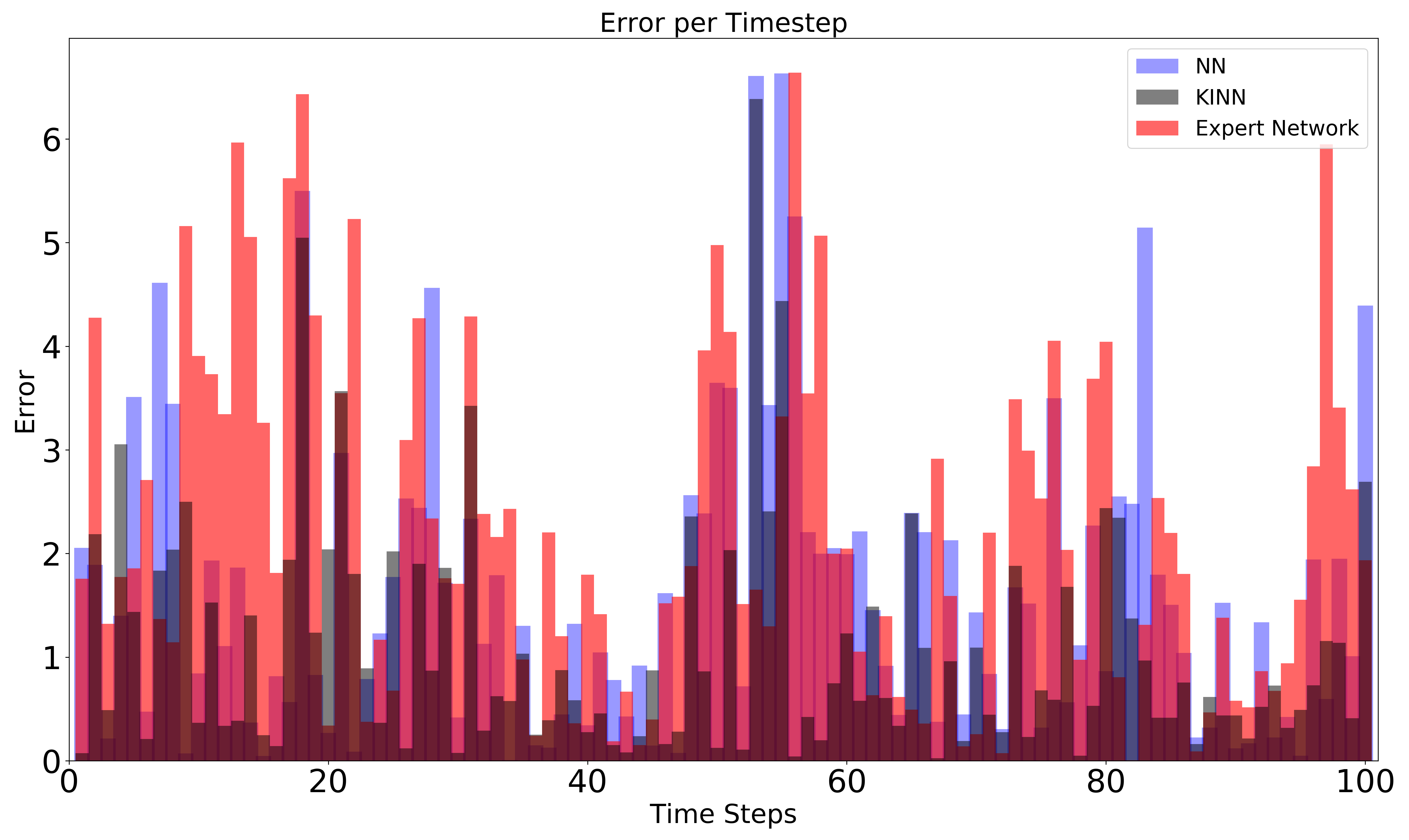}}
    \caption{Prediction and error plot with inaccurate expert prediction}
    \label{fig:noisy_preds}
\end{figure*}

In all of the previous experiments, the expert model was relatively better compared to the LSTM model employed in our experiments. The obtained results highlights KINN's ability to capitalize over the information obtained from the expert model to achieve significant improvements in its prediction. 
KINN also demonstrated amazing generalization despite of drastic reduction in the amount of training data, highlighting KINN's ability to achieve accurate predictions in low data regimes. 
However, in conjunction to reducing dependency of the network on data, it is also imperative that the network does not become too dependent on the expert knowledge making it essential to be accurate/perfect. This is usually not catered for in most of the prior work. 
%The residual learning scheme equipped the system with the ability to automatically adjust it's predictions according to the quality of information contained in predictions made by the expert network. This adaptability enabled KINN to decide its intervention in the prediction made by the expert.
We believe that the proposed residual scheme enabled the network to handle erroneous expert knowledge efficiently by allowing it to be smart enough to realize weaknesses in the expert network and adjust accordingly. 
In order to verify KINN's ability to adjust with poor predictions from the expert, we performed another experiment where random noise was injected into the predictions from the expert network. This random noise degraded the reliability of the expert predictions.
% Hence it will only be fair if the proposed network is tested in a scenario where expert information is comparatively inaccurate. 
To achieve this, random noise within one standard deviation of the average traffic flow was added to the expert predictions. As a result, the resulting expert predictions attained a MSE of 7.81 which is considerably poor compared to that of the LSTM (5.90). We then trained KINN using these noisy expert predictions. 
% and whole of the training set is used for training the network. 
Fig.~\ref{fig:noisy_preds} visualizes the corresponding prediction and error plots.  

As evident from Fig.~\ref{fig:noisy_pred}, KINN still outperformed both the expert as well as the LSTM with a MSE of 3.09. 
Despite the fact that neither the LSTM, nor the expert model was accurate, KINN still managed to squeeze out useful information from both modalities to construct an accurate predictor. 
% Although none of the baseline models were relatively accurate enough the proposed network still managed to utilize useful information in both of the modalities and was able to combine it constructively to produce a better prediction. 
%This validates KINN's ability to overcome inaccuracies in the expert predictions in an automated fashion. 
%based on its authenticity in order to come up with reliable predictions.
This demonstrates true strength of KITNN as it not only reduces dependency of the network on the data but also adapts itself in case of poorly made expert opinions. 
KINN achieved a significant reduction of 48\% in the MSE of the LSTM network by incorporating the noisy expert prediction in the residual learning framework.
% The results are quite encouraging since for this particular experimental setting there was 52\% reduction in MSE compared to NN baseline model, which was better among the two baseline models. 

\subsection{Experiment \# 04: Reduced training set and noisy expert} \label{sec:noisyExpReducedData}

\begin{figure*}[!t]
    \centering
    \subfigure[Predictions of all models]{\includegraphics[width=0.48\textwidth]{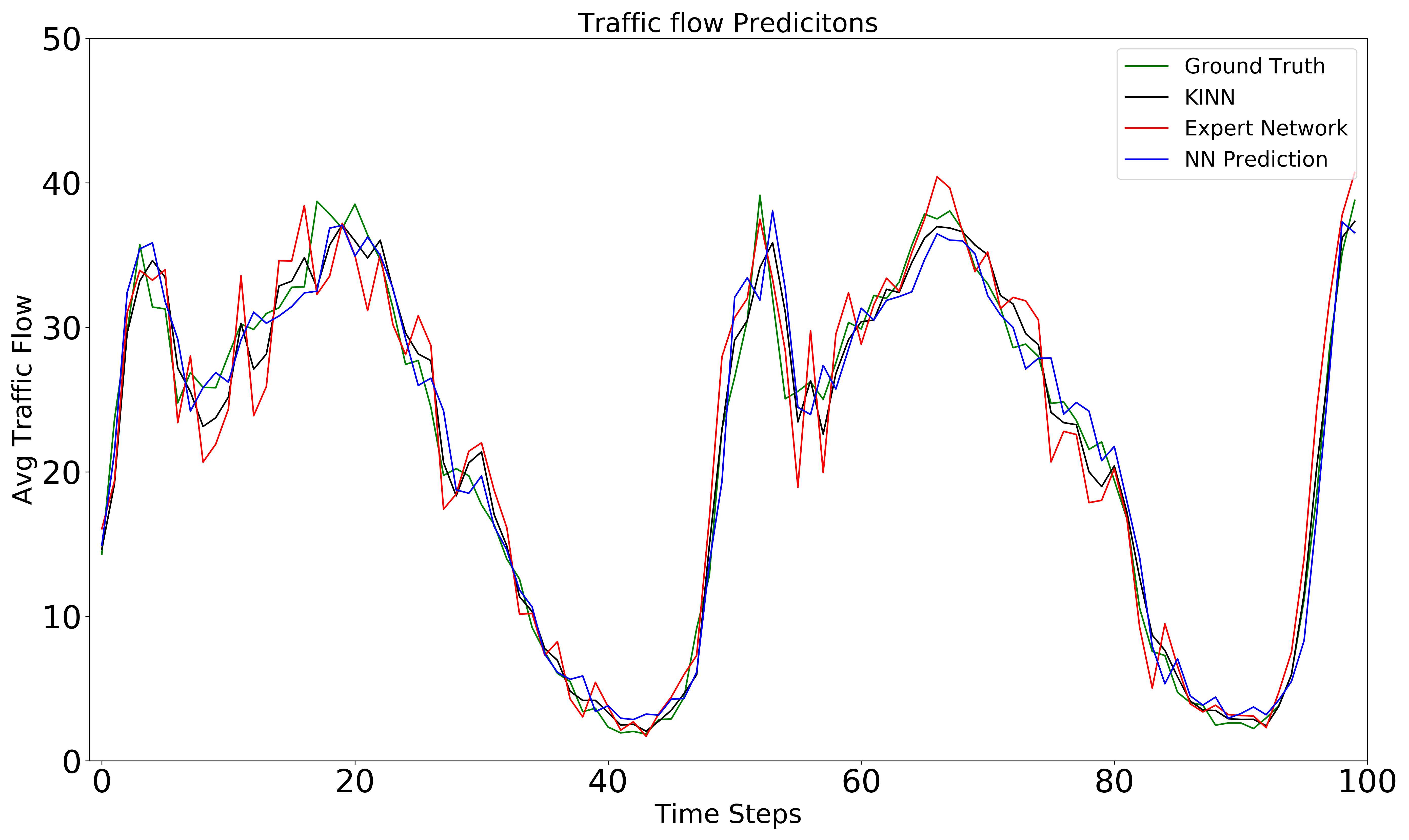}\label{fig:noisy_10pred}}
    \subfigure[Step wise error plot]{\includegraphics[width=0.48\textwidth]{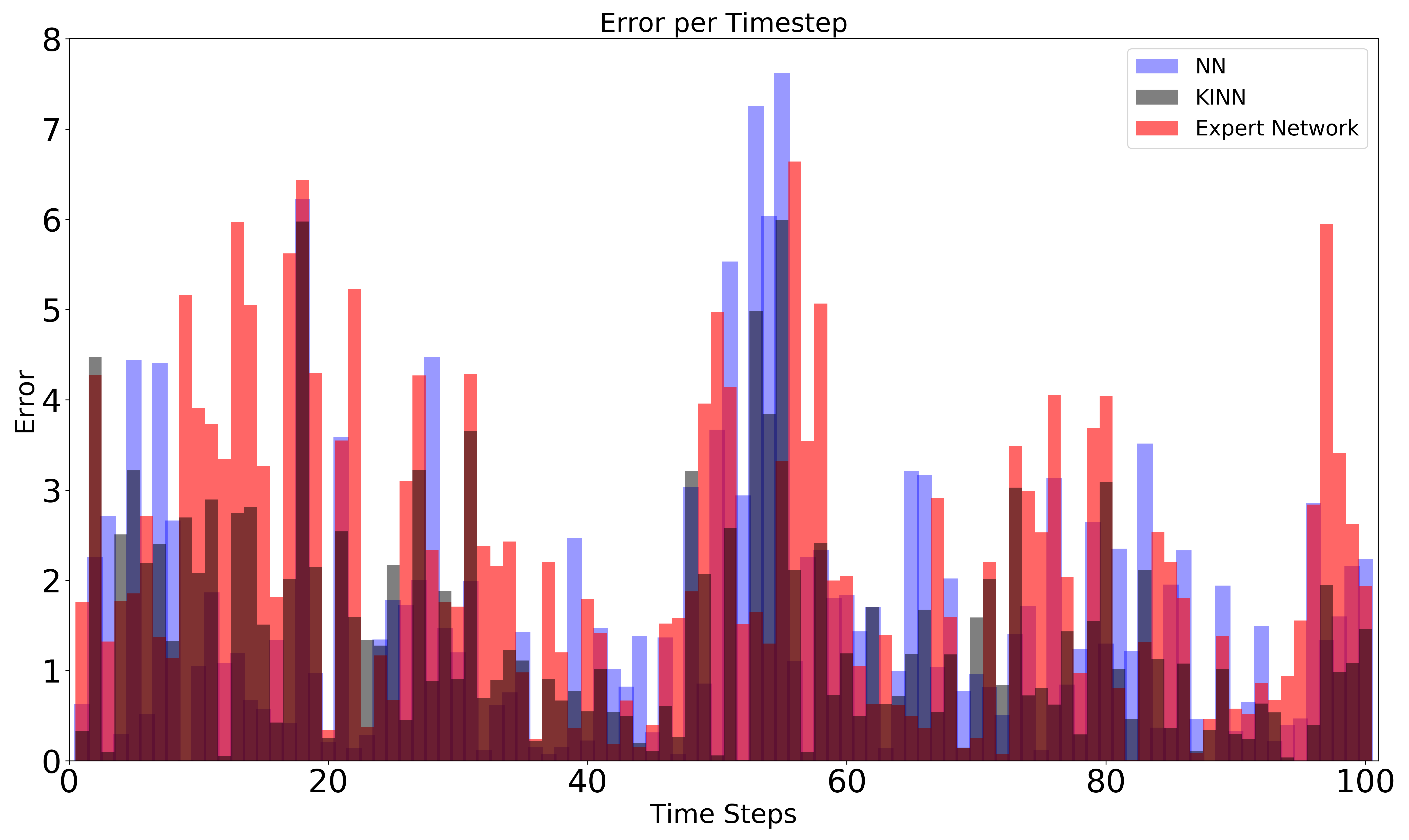}}
    \caption{Prediction and error plot with inaccurate expert prediction and with only 10\% data}
    \label{fig:noisy_10preds}
\end{figure*}

As a natural followup to the last two experiments, we introduced both conditions at the same time i.e. reduced training set size and noisy predictions from the expert.
% As a final experiment, the performance of KINN as also evaluated in worst of the scenarios, where we simultaneously regarded both the amount of training data as well as quality of the expert predictions. 
The training set was again reduced to 10\% subset of the training data for training the model while keeping the testing set intact. Fig.~\ref{fig:noisy_10preds} demonstrates that despite this worst condition, KINN still managed to outperform both the LSTM as well as the noisy expert predictions. 

\subsection{Experiment \# 05: Full training set and poor expert} \label{sec:noExpFullData}

As the final experiment, we evaluated KINN's performance in cases where the expert predictions are not useful at all. We achieved this via two different settings. In the first setting, we considered that the expert network predicts zero every time. In the second setting the expert network was made to lag by a step of one resulting in mismatch of the time step with the predictions.
%This setup is in term of equivalent to the the unconditioned case where the full projection has to be learned. 
Putting zero in place of $\hat{x}^{p}_{t}$ in Eq.~\ref{conditionedOptimProblem} yields:

\begin{equation*}
     \hat{x}_{t} = \phi([x_{t-1}, x_{t-2},..., x_{t-p}, 0]; \mathcal{W}) + 0
 \end{equation*}

\begin{equation*}\label{conditionedOptimProblemWithNoExp}
\begin{aligned}
      \mathcal{W}^{*} = \argmin{\mathcal{W}} \frac{1}{|\mathcal{X}|} \sum_{\mathbf{x} \in \mathcal{X}} (x_{t} - (\phi([x_{t-1},..., x_{t-p}, 0]; \mathcal{W}) \\+ 0))^2 \\
      \mathcal{W}^{*} = \argmin{\mathcal{W}} \frac{1}{|\mathcal{X}|} \sum_{\mathbf{x} \in \mathcal{X}} (x_{t} - (\phi([x_{t-1},..., x_{t-p}, 0]; \mathcal{W}))^2
\end{aligned}
\end{equation*}

This is almost equivalent to the normal unconditioned full input to output space projection learning case (Eq.~\ref{unconditionedOptimProblem}) except a zero in the conditioning vector. However, in case of lagged predictions by the expert network, since we stack the expert prediction $\hat{x}^{p}_{t}$ in a separate channel, the network assigns a negligible weight to this channel, resulting in exactly the same performance as the normal case.

Table~\ref{table:1} provides the details regarding the results obtained for this experiment. It is clear from the table that in cases where the expert network either gave zero as its predictions or gave lagged predictions, which is useless, the network performance was identical to the normal case since the network learned to ignore the output from the expert. 
These results highlight that KINN provides a lower bound on the performance based on the performance of the two involved entities: expert model and the network.

\subsection{Discussion}

These thorough experiments advocates that the underlying residual mapping function learned by KINN is successful in combining the network with the prediction made by the expert. Specifically, KINN demonstrated the ability to recognize the quality of the prediction made by both of the base networks and shifted its reliance according to it. In all of the experiments that we have conducted, MSE of the predictions made by KINN never exceeded (disregarding insignificant changes) the MSE of the predictions achieved by the best among the LSTM and the expert model except in case of completely useless expert predictions, where it performed on par with the LSTM network. Table~\ref{table:1} provides a summary of the results obtained from all the different experiments performed. It is interesting to note that even with a huge reduction in the size of the training set, the MSE does not drastically increase as one would expect. This is due to the strong seasonal component present in the dataset. As a result, even with only 10\% subset of the training data, the algorithms were able to learn the general pattern exhibited by the sequence. It is only in estimating small variations that these networks faced difficulty when training on less data.

% SAS: Future work
%\textcolor{red}{Future work ...}
%A promising direction for the future is to extend this residual learning framework for classification settings. We intend to rigorously test the proposed method on a range of different settings and datasets in the future.

\section{Conclusion}

We propose a new architecture for incorporating expert knowledge into the deep network. It incorporates this expert knowledge in a residual scheme where the network learns a correction term for the predictions made by the expert. 
% The proposed architecture alleviates shortcomings of the traditional NN of being too data dependent. 
The knowledge incorporation scheme introduced by KINN has three key advantages. The first advantage is regarding the relaxation of the requirement for a huge dataset to train the model. The second advantage is regarding the provision of a lower bound on the performance of the resulting classifier since KINN achieves the best of both worlds by combining the two different modalities. The third advantage is its robustness in catering for poor/noisy predictions made by the expert.
Through extensive evaluation, we demonstrated that the underlying residual function learned by the network makes the system robust enough to deal with imprecise expert information even in cases where there is a dearth of labelled data. 
This is because the network does not try to imitate predictions made by the expert network, but instead extracts and combines useful information contained in both of the domains. 

\section{Acknowledgements}
This work is supported by Higher Education Commission (Pakistan), Continental and BMBF project DeFuseNN (Grant 01IW17002).

\bibliography{bibfile.bib} 
\bibliographystyle{aaai}
\end{document}